# A Novel Online Multi-label Classifier for High-Speed Streaming Data Applications

Rajasekar Venkatesan, Meng Joo Er, Mihika Dave, Mahardhika Pratama, Shiqian Wu

**Abstract** – In this paper, a high-speed online neural network classifier based on extreme learning machines for multi-label classification is proposed. In multi-label classification, each of the input data sample belongs to one or more than one of the target labels. The traditional binary and multi-class classification where each sample belongs to only one target class forms the subset of multi-label classification. Multi-label classification problems are far more complex than binary and multi-class classification problems, as both the number of target labels and each of the target labels corresponding to each of the input samples are to be identified. The proposed work exploits the high-speed nature of the extreme learning machines to achieve real-time multi-label classification of streaming data. A new threshold-based online sequential learning algorithm is proposed for high speed and streaming data classification of multi-label problems. The proposed method is experimented with six different datasets from different application domains such as multimedia, text, and biology. The hamming loss, accuracy, training time and testing time of the proposed technique is compared with nine different state-of-the-art methods. Experimental studies shows that the proposed technique outperforms the existing multi-label classifiers in terms of performance and speed.

## 1. Introduction

Classification is a problem of identifying which of the target categories a data sample belongs to. In machine learning, classification can be defined as "Given a set of training examples composed of pairs, find a function f(x) that maps each attribute vector $x_i$ to its associated class $y_i$, i = 1,2,….,n, where n is the total number of training samples" (de Carvalho and Freitas 2009). This is the most common type of classification problem called single-label classification. In single-label classification, each of the data sample belongs to only one of the target labels. But in real world applications, there may be several cases in which each data sample belongs to more than one target labels. This results in the need for multi-label classification. The multi-label classification problems have gained much importance due to its rapidly increasing application areas. The application areas of multi-label classification include but are not limited to text categorization (Gonçalves and Quaresma 2003; Joachims 1998; Luo and Zincir-Heywood 2005; Tikk and Biró 2003; Yu et al. 2005), bioinformatics (Elisseeff and Weston 2001; Min-Ling and Zhi-Hua 2005), medical diagnosis (Karali and Pirnat 1991), image/scene and video categorization (Boutell et al. 2003; Shen et al. 2003), genomics, map labeling (Zhu and Poon 1999), marketing, multimedia, emotion, music categorization, etc. In recent years, the multi-label classification has drawn increased research attention due to the realization of the omnipresence of multi-label prediction tasks in several areas (Tsoumakas et al. 2010). Due to the wide range of applications and increasing importance, several multi-label classification techniques have been developed and are available in the literature.

The learning techniques in machine learning can be grouped into two broad categories: Batch Learning and Online Learning. In batch learning, all the training data are collected in prior, and the parameters of the network are calculated by processing all the training data concurrently. This poses a major limitation on the batch learning techniques as they are unable to learn from streaming data. On the other hand, in online/sequential learning techniques the network parameters are updated iteratively with single-pass learning procedure (Pratama et al. 2015a; Pratama et al. 2015d). Several books (Angelov 2012; Gama 2010; Kasabov 2007; Sayed-Mouchaweh and Lughofer 2012) are available in the literature that comprehensively elaborates the data stream classification. In many cases, online learning is preferred over batch learning as they can learn from data streams (Pratama et al. 2015b; Pratama et al. 2015c)and do not require re-training whenever a new data sample is received.

As foreshadowed, in single-label classification problems, each of the sample data is associated with a unique target class label from a pool of target class labels. Single-label classifiers can be further classified into binary classifiers and multi-class classifiers. Binary classification is the most trivial classification problem in which the input sample belongs to one of the two target class labels. Medical diagnosis, biometric security, and other similar applications are examples of binary classification. When the total number of target class labels is greater than two, it is called multi-class classification. In multi-class classification, each of the input samples corresponds to a unique class among a pool of target class labels. Character recognition (Mohiuddin and Mao 2014), biometric identification (Song et al. 2013; Srivastava et al. 2015), and other related applications are examples of multi-class classification. Several online machine learning classifiers for single-label classification are available in literature (Lughofer and Buchtala 2013; Polikar et al. 2001). Evolving classifiers (Bouchachia 2010; Iglesias et al. 2010) and fuzzy systems based classifiers (Angelov et al. 2008; Lemos et al. 2013; Xydeas et al. 2006) have also been developed for streaming data applications. However, there are several real-world classification problems in which the target labels are not mutually exclusive, and each of the data samples corresponds to more than one target labels resulting in need for multi-label classification. The traditional binary and multi-class classification are special cases of multi-label classification problems. Thus, being the superset of binary and multi-class classification problems, it can be stated that the multi-label classification forms the generalization of all classification problems. Due to its generality, the multi-label classification problems are more difficult and more complex when compared to single-label classification problems (Zhang and Zhou 2007).

Several approaches for solving multi-label problems are available in the literature. But most of the available approaches are based on batch learning techniques. Online techniques for multi-label classification are still greatly to be explored. The paper on streaming multi-label classification by Reed and his team (Read et al. 2011) list out the existing classifiers on multi-label classification for streaming applications. The existing techniques listed belongs to the category of problem transformation methods. In problem transformation methods, the multi-label classification problem is converted into multiple single-label classification problem and it uses existing single-label techniques for classification. The proposed method, on the other hand extends the base algorithm itself to adapt to the multi-label problems. Therefore, the proposed method differs significantly from the existing problem transformation based techniques. It is also to be highlighted that the proposed method is the first extreme learning machine based real-time online multi-label classifier. The proposed method employs a new threshold-based classification for multi-label problems. Unlike single-label classification, the number of target labels differs for every data sample. Therefore, in multi-label classification, both the number of labels and the corresponding labels are unknown. Also, different multi-label datasets differs significantly from each other with respect to label density and label cardinality characteristics. A classifier that performs well in one dataset may not necessarily perform well in a different dataset. Due to the increased complexity of the multi-label classification caused by its generality, the time taken for training the classifier is high for most of the techniques. Also, the highly complex nature of the multi-label classification problems poses a considerable challenge in developing high-speed real-time online classifiers. The application areas of multi-label classification are increasing rapidly due to its generality. Several real world applications require the need for multi-label classification. High-speed processing of streaming data for multi-label classification is highly essential for real-world real-time applications. The proposed work exploits the high-speed nature of extreme learning machines, and a novel online multi-label classifier is developed. The proposed ELM based online multi-label classifier outperforms the existing multi-label classifiers in performance and speed.

The rest of the paper is organized as follows. A condensed overview of multi-label classification and different types of multi-label classifiers are discussed in Section 2. Details of the proposed approach are described in Section 3. Section 4 describes the experimental specifications and the different benchmark metrics used for analyzing the multi-label classification datasets. The performance of the proposed method, performance comparison with existing methods and related discussions are carried out in Section 5. Finally, concluding remarks are given in Section 6.

## 2. Multi-label Classifier

In single-label classifications such as binary and multi-class classification, the target labels of each sample are unique and the target labels are mutually exclusive, i.e. Consider there are M target classes, and $p_i$ denotes the probability that the input sample is assigned to $i^{th}$ class. Then, for single-label classification the following equality condition holds true.

$$\sum p_i = 1 \qquad (1)$$

On contrary, this equality does not hold true for multi-label problems. Also, from (Elisseeff and Weston 2001), it can be seen that the binary classification problem, multi-class classification problems and ordinal regression problems are special cases of the multi-label problems with the number of target labels corresponding to each of the data sample restricted to 1. The definition of multi-label learning as given by (Sorower 2010) is, "Given a training set, $S = (x_i, y_i)$, $1 \leq i \leq n$, consisting of n training instances, ($x_i \in X$, $y_i \in Y$) drawn from an unknown distribution D, the goal of multi-label learning is to produce a multi-label classifier h:X→Y that optimizes some specific evaluation function or loss function".

The problem of multi-label learning can be summarized as follows:

— There exists an input space X of feature dimension D. $x_i \in X$, $x_i = (x_{i1}, x_{i2}, \ldots x_{iD})$
— There exists a label space L of dimension M. $L = \{\zeta_1, \zeta_2, \ldots, \zeta_M\}$
— Consider there are N training samples, each of the training samples can be represented by a pair of tuples (input space and label space). $\{(x_i, y_i) \mid x_i \in X, y_i \in Y, Y \subseteq L, 1 \leq i \leq N\}$
— A training model that maps the input tuple to output tuple.

There are several multi-label classifiers available in the literature. The existing techniques can be broadly classified into two categories: Batch learning techniques and online learning techniques. The batch learning based multi-label classifiers are further classified by (Madjarov et al. 2012; Tsoumakas and Katakis 2006) into Problem Transformation (PT) methods, Algorithm Adaptation (AA) methods and Ensemble (EN) methods. There are very limited number of online multi-label classifiers available in the literature. A brief summary of existing multi-label classifiers is discussed in this section. An overview of multi-label methods is shown in Fig. 1.

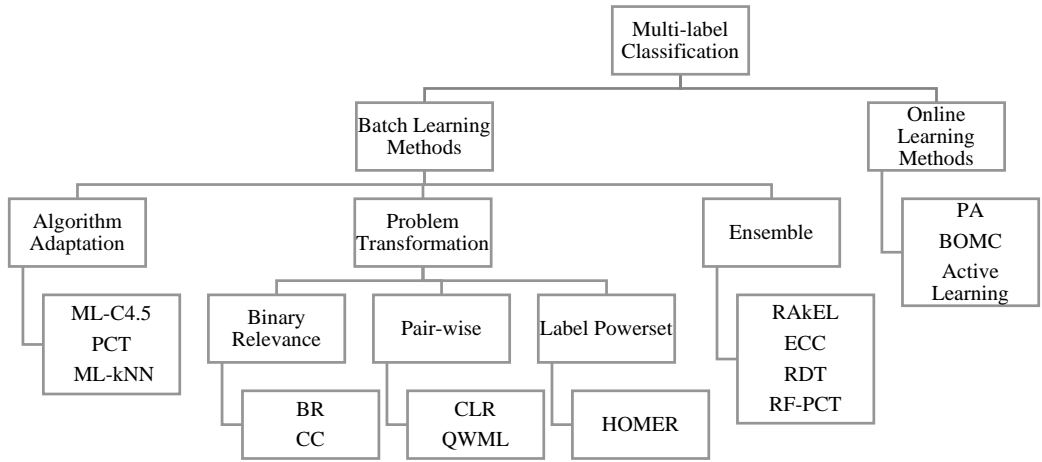

Fig. 1. Overview of multi-label methods

## 2.1 Batch Learning Methods

There are several batch learning based multi-label classification techniques available in the literature. (Tsoumakas and Katakis 2006) categorized the techniques into two categories: PT methods and AA methods. Later, (Madjarov et al. 2012) extended the classification to include a third category of methods: EN methods.

### 2.1.1 Problem Transformation (PT) methods

PT methods, as the name implies, transform the multi-label classification problems into multiple single-label classification problems and employ existing single-label classifiers to perform the classification and finally combines the individual classifier results to provide the multi-label classification results. The PT methods are further divided into three categories: Binary relevance methods (Binary Relevance (BR), Classifier Chaining (CC)), Pairwise methods (Calibrated Label Ranking (CLR), Qweighted multi-label (QWML) approach) and Label powerset method (HOMER).

### 2.1.2 Algorithm Adaptation (AA) Methods

In AA methods, the base algorithm corresponding to the classification is itself extended to adapt to multi-label problems. AA methods are algorithm-dependent methods. Several AA methods for multi-label classification are available such as Predictive Clustering Trees (PCTs), Multi-label k-nearest neighbors (ML-kNN), ML-C4.5, etc. Techniques like SVM, neural networks, Boosting also have multi-label variants.

### 2.1.3 Ensemble (EN) methods

EN methods employ an ensemble of PT or AA methods. Ensemble Classifier Chains (ECC) is an EN method that uses CC as base technique and forms an ensemble of multiple CC methods to address the multi-label problems. Techniques like PCT, decision trees (DT), ML-C4.5 are used in ensemble with Random Forest (RF) to form RF-PCT, RDT, and RFML-C4.5 respectively. Random-k label sets (RAkEL) uses label power set for classifying each of the label sets.

## 2.2 Online Methods

Due to the complicated nature of multi-label problems, very few works are available in online learning for multi-label classification. Some of the significant works are briefly reviewed. (Crammer 2004) proposed Passive-Aggressive (PA) method for multi-label classification. In the year 2010, (Zhang et al. 2010) proposed a method called Bayesian Online Multi-label Classification (BOMC). Due to its various real life applications, Microsoft focused its research on multi-label classification and developed an online multi-label active learning technique for multimedia applications (Hua and Qi 2008). From the lack of mentioning of any online multi-label classification methods in any of the multi-label review articles thus far, it is evident that there are no generic online multi-label classification techniques that can be applied to a wide range of application domains. The PA and the BOMC techniques are implemented only for text categorization datasets and the Active Learning framework from Microsoft is application specific to multimedia datasets.

This paper proposes an ELM based online multi-label classifier that is capable of performing online multi-label classification on streaming data in real-time. There are no online multi-label techniques available in the literature which can perform real-time multi-label classification. The proposed technique is experimented on six datasets from different application domains.

## 3. Proposed Approach

ELM is a single-hidden layer feedforward neural network based learning technique (Ding et al. 2015; Huang et al. 2006). A key feature of ELM is that it maintains the universal approximation capability of single hidden-layer feedforward neural network. It has gained much attention due to its special nature of random input weight initialization and its unique advantage of extreme learning speed (Wang et al. 2011).In ELM, the initial weights and the hidden layer bias can be selected at random, and the network can be trained for the output weights to perform the classification (Huang 2015; Huang et al. 2011; Ning et al. 2014; Ning et al. 2015; Wang et al. 2014; Wang et al. 2015). This results in a fast learning speed and generalization of performance. The proposed method exploits these advantages of the ELM for online multi-label classification.

The pre-processing and post-processing of data is of prime importance in extending ELM technique for online multi-label problems. As opposed to single-label classification in which each of the input samples belongs to only one of the target labels, in multi-label problems, each input sample may belong to one or more samples. Therefore, the classifier should be able to predict both the number of labels an input sample belongs to and each of the target labels that corresponds to the input sample. It is also to be noted that, not all multi-label datasets are equally multi-labelled. The degree of multi-labelness varies among different datasets and different applications. This results in increased complexity of the multi-label problem resulting in much longer training and testing time of the multi-label classification technique. A brief review on ELM and Online Sequential ELM (OS-ELM) is presented to provide basic background information

## 3.1 Extreme Learning Machines

Consider there are N training samples of the form $\{(x_i,y_i)\}$, $x_i = [x_{i1},x_{i2},...,x_{in}]^T \in R^n$ and $y_i = [y_{i1},y_{i2},...y_{im}]^T$. In multi-label classification, each input sample belongs to a subset of labels from the label space given as $Y \subseteq L$, $L = \{\zeta_1, \zeta_2, ...., \zeta_m\}$. Let $\bar{N}$ be the number of hidden layer neurons, the output 'o' of the SLFN is given by

$$\sum_{i=1}^{\bar{N}} \beta_i g_i(x_j) = \sum_{i=1}^{\bar{N}} \beta_i g(w_i \cdot x_j + b_i) = o_j \qquad (2)$$

where, $\beta_i = [\beta_{i1},\beta_{i2},...\beta_{im}]^T$ is the output weight, $g(x)$ is the activation function, $w_i = [w_{i1},w_{i2},...w_{in}]^T$ is the input weight and $b_i$ is the hidden layer bias.

The input weights $w_i$ and the hidden layer bias $b_i$ are randomly assigned in ELM. Therefore, the network must be trained for $\beta_i$ such that the output of the network is equal to the target class so that the error difference between the actual output and the predicted output is 0.

$$\sum_{j=1}^{\bar{N}} \|o_j - y_j\| = 0 \qquad (3)$$

Thus, the ELM classifier output can be as follows:

$$\sum_{i=1}^{\bar{N}} \beta_i g(w_i \cdot x_j + b_i) = y_j \qquad (4)$$

The above equation can be written in following matrix form:

$$H\beta = Y \qquad (5)$$

where,

$$H = \begin{bmatrix} g(w_1 \cdot x_1 + b_1) & \cdots & g(w_{\bar{N}} \cdot x_1 + b_{\bar{N}}) \\ \vdots & \ddots & \vdots \\ g(w_1 \cdot x_N + b_1) & \cdots & g(w_{\bar{N}} \cdot x_N + b_{\bar{N}}) \end{bmatrix}_{N \times \bar{N}} \quad (6)$$

$$\beta = \begin{bmatrix} \beta_1^T \\ \vdots \\ \beta_{\bar{N}}^T \end{bmatrix}_{\bar{N} \times m} \quad (7)$$

$$Y = \begin{bmatrix} y_1^T \\ \vdots \\ y_N^T \end{bmatrix}_{N \times m} \quad (8)$$

The output weights of the ELM network can be estimated using the equation

$$\beta = H^+ Y \quad (9)$$

where $H^+$ is the Moore-Penrose inverse of the hidden layer output matrix H, and it can be calculated as follows:

$$H^+ = (H^T H)^{-1} H^T \quad (10)$$

The theory and mathematics behind the ELM have been extensively discussed in (Ding et al. 2015; Huang 2015; Huang et al. 2011; Huang et al. 2006) and hence are not re-stated here.

### 3.2 The Proposed OSML-ELM

The various steps involved in the proposed method are briefly stated. The key novelty of the proposed method is that, there are no online multi-label classification techniques available in the literature that can perform classification in real-time on streaming data. The proposed method is the multi-label formulation of the online sequential extreme learning machine and hence called Online Sequential Multi-label ELM (OSML-ELM).

*Initialization of Parameters.* Fundamental parameters such as the number of hidden layer neurons and the activation function are initialized. Sigmoidal activation function is used for the experimentation. The problem of overfitting is tackled by using the early stopping technique. In early stopping technique, the point at which the training accuracy increases at the expense of generalization error is identified and further training is stopped. The number of hidden neurons are selected depending upon the nature and complexity of the dataset while preventing the overfitting of data.

*Processing of Inputs.* In traditional single-label problems, the target class will be a single-label associated with the input sample. But, in the multi-label case, each input sample can be associated with more than one class labels. Hence, each of the input samples will have the associated output label as an m-tuple with 0 or 1 representing the belongingness to each of the labels in the label space L. This is a key difference between the inputs available for single-label and multi-label problems. As opposed to single-label classification with a single target label, the multi-label problem has a target label set which is a subset of label space L. The label set denoting the belongingness for each of the labels is converted from unipolar representation to bipolar representation.

*ELM Training.* The processed input is then supplied to the online sequential variant of ELM technique. Let H be the hidden layer output matrix, β be the output weights and Y be the target label, the ELM can be represented in a compact form as Hβ = Y where Y⊆L, L = {ζ₁, ζ₂,...., ζₘ}. During the training phase, Let $N_0$ be the number of samples in the initial block of data that is provided to the network. The initial output weight $β_0$ is calculated from equation 9 and 10.

β = H⁺Y and H⁺ = (HᵀH)⁻¹Hᵀ,

Consider **M₀ = (H₀ᵀH₀)⁻¹**, therefore, **β₀ = M₀H₀ᵀY₀**.

For each of the subsequent sequentially arriving data, the output weights can be updated by incorporating the recursive least square algorithm with the ELM learning as

$$M_{k+1} = M_k - \frac{M_k h_{k+1} h_{k+1}^T M_k}{1 + h_{k+1}^T M_k h_{k+1}} \tag{11}$$

$$\boldsymbol{\beta}_{k+1} = \boldsymbol{\beta}_k + M_{k+1} h_{k+1}(Y_{k+1}^T - h_{k+1}^T \boldsymbol{\beta}_k) \tag{12}$$

where k = 0,1,2.... N-N₀-1.

The detailed mathematics and derivation behind the recursive least square based online ELM learning called online-sequential extreme learning machine is discussed in detail in several literatures (Li et al. 2007; Liang et al. 2006).

*ELM Testing.* In the testing phase, the test data sample is evaluated using the values of β obtained during the training phase. The input data that can be a combination of Boolean, discrete and continuous data type is given to the network. The network then computes Y = Hβ. The predicted output Y obtained is a set of real numbers of dimension equal to the number of labels.

*Post-processing and Multi-label Identification.* The prime step in extending the ELM based technique for online multi-label problems is the post-processing and thresholding. In binary and multi-class classification, each of the input sample belongs to only one target label and, therefore, can be identified as the index of the maximum value in the predicted output. On contrary, in multi-label classification the number of labels each sample belongs to is not constant. Each input sample may belong to one or more than one of the target labels. Therefore, the classifier must predict both the number of labels and each of the corresponding labels for the input data sample. The number of labels corresponding to a data sample is completely unknown. Hence, in the proposed method, a thresholding-based label association is proposed. The threshold value is selected during the training phase such that it maximizes the separation between the family of labels the input belongs to and the family of labels the input does not belong to, based on the raw output values Y. Setting up of the threshold value is of prime importance as it directly affects the performance of the classifier. The L dimensioned raw-predicted output is compared with a unique threshold value. The index values of the predicted output Y which are greater than the fixed threshold represent the belongingness of the input sample to the corresponding class.

Setting the threshold value is of critical importance. The threshold value is selected such that it maximizes the difference between the category of labels to which the sample belongs to and the category of labels to which the sample does not belong to with respect to the raw output values Y obtained during the training phase. The distribution of the raw output values of Y for categories of labels that the input sample belongs to ($Y_A$) and the categories of labels the input sample does not belong to ($Y_B$) are identified. Based on the distribution of $Y_A$ and $Y_B$, a threshold value is identified using the formula,

$$\text{Threshold value} = (\min(Y_A) + \max(Y_B))/2 \tag{13}$$

As a trivial case, the threshold can be set as zero. In which case, the raw predicted output values will be passed as arguments to a bipolar step function. The threshold value is compared with the raw output values of Y estimated by the classifier, and the number of target labels corresponding to the data sample is identified. Then, based on the threshold value, the subset of labels that corresponds to the input data sample is recognized. The threshold value is determined by analyzing the distribution of the raw predicted output values during the training phase. From the distribution, a particular value is chosen that maximizes the separation between the two categories of the labels. The proposed method belongs to the category of algorithm adaptation method, where the base algorithm is adapted to perform multi-label classification problems. It is to be highlighted that there are no ELM-based online multi-label classifiers in the literature thus far. The proposed method is the first to adapt the ELM for online multi-label problems and make extensive experimentation, results comparison and analysis with the state-of-the-art techniques. The overview of the proposed algorithm is summarized.

---

**Algorithm: Proposed OSML-ELM algorithm for multi-label classification**

1. The parameters of the network are initialized

2. The raw input data is processed for classification

3. ELM Training – Initial phase
   Processing of initial block of data
   $M_0 = (H_0^T H_0)^{-1}$
   $\beta_0 = M_0 H_0^T Y_0$

4. ELM Training – Sequential phase
   Online processing of sequential data

   $$M_{k+1} = M_k - \frac{M_k h_{k+1} h_{k+1}^T M_k}{1 + h_{k+1}^T M_k h_{k+1}}$$

   $$\beta_{k+1} = \beta_k + M_{k+1} h_{k+1} (Y_{k+1}^T - h_{k+1}^T \beta_k)$$

5. ELM Testing
   Estimation of raw output values using $Y = H\beta$

6. Thresholding
   Applying the threshold value based on separation between two categories of labels ($Y_A$ and $Y_B$). Threshold value = $(\min(Y_A) + \max(Y_B))/2$
   Identifying the number of labels corresponding to input data sample
   Identifying the target class labels for the input data sample

---

## 4. Experimentation

This section describes the different multi-label dataset metrics and gives the experimental design used to evaluate the proposed method.

Multi-label datasets have a unique property called the degree of multi-labelness. In order to quantitatively measure the multi-labelness of a dataset, two dataset metrics are available in the literature. They are Label Cardinality (LC) and Label Density (LD). Not all datasets are equally multi-labelled. The number of labels, the number of samples having multiple labels, the average number of labels corresponding to a particular sample varies among different datasets resulting in a varied degree of multi-labelness to a dataset.

Consider there are N training samples and the dataset is of the form $\{(x_i, y_i)\}$ where $x_i$ is the input data and $y_i$ is the target label set. The target label set is a subset of labels from the label space L of dimensionality m, given as $Y \subseteq L$, $L = \{\zeta_1, \zeta_2 \ldots \zeta_m\}$.

Definition 4.1 (Tsoumakas and Katakis 2006) Label Cardinality of the dataset is the average number of labels of the examples in the dataset.

$$Label - Cardinality = \frac{1}{N} \sum_{i=1}^{N} |Y_i| \qquad (14)$$

Label Cardinality is independent of the number of labels present in the dataset and signifies the average number of labels present in the dataset.

Definition 4.2 (Tsoumakas and Katakis 2006) Label Density of the dataset is the average number of labels of the examples in the dataset divided by |m| where m is the dimensionality of label set L.

$$Label - Density = \frac{1}{N} \sum_{i=1}^{N} \frac{|Y_i|}{|m|} \qquad (15)$$

Label density takes into consideration the number of labels present in the dataset. Bernardini et al. (2014) analyzed the effect of label density and label cardinality on multi-label learning. It is to be noted that, two datasets with same label cardinality but different label density can significantly vary and may result in different behavior of the training algorithm (Zhang and Zhou 2007). The influence of label density and label cardinality on multi-label learning is analyzed by (Bernardini et al. 2014).

The proposed method is experimented with five benchmark datasets comprising of different application areas such as multimedia, text, and biology. The performance of the proposed method is compared with that of 9 existing methods from batch learning and 1 method from online learning method. The proposed method is experimented with datasets from different application domains and exhibit wide range of label density and label cardinality. The number of target class labels ranges from 6 labels to 374 labels, and the number of features or attributes in the dataset ranges from 103 to 1449. The dataset metrics such as label cardinality varies from as low as 1.07 to as high as 4.24. Label cardinality of 1.07 represents that each of the input samples corresponds to 1.07 labels on average. Label cardinality of 4.24 signifies that each sample on an average corresponds to 4.24 labels. Since the label density is inversely proportional to the number of labels present in the dataset, lower the label density value indicates that only fewer samples correspond to a particular label, thus, posing a challenge for the multi-label techniques to train fast enough so as to learn the target label set within the limited samples. The specifications of the datasets are given in Table 1. The number of samples in each of the dataset used for training and testing phase and the feature dimension are included in the dataset specifications. The datasets are obtained from KEEL multi-label dataset repository.

Table 1. Dataset specifications

| Dataset | Domain | No. of Features | No. of Samples | #Train | #Test | No. of Labels | LC | LD |
|---|---|---|---|---|---|---|---|---|
| Yeast | Biology | 103 | 2417 | 1600 | 817 | 14 | 4.24 | 0.303 |
| Scene | Multimedia | 294 | 2407 | 2000 | 407 | 6 | 1.07 | 0.178 |
| Corel5k | Multimedia | 499 | 5000 | 4500 | 500 | 374 | 3.52 | 0.009 |
| Enron | Text | 1001 | 1702 | 1200 | 502 | 53 | 3.38 | 0.064 |
| Medical | Text | 1449 | 978 | 700 | 278 | 45 | 1.25 | 0.027 |

The hamming loss, training and testing time of the proposed method are compared to 9 different multi-label techniques available in the literature. The 9 techniques are chosen such that they are from PT, AA and EN methods. The implementation procedure of all the 9 techniques are adapted from the extensive experimental comparison work on multi-label classifiers by Madjarov and team (Madjarov et al. 2012). Also, the chosen techniques belong to different learning paradigms such as SVM, decision trees, and nearest neighbors. The details of state-of-the-arts multi-label techniques used for result comparison are given in Table 2.

## 5. Results and Discussions

The proposed method is experimented with each of the datasets mentioned in Table 3 and is compared with 9 state-of-the-art multi-label classification techniques. Also, the performance of the proposed method is compared with the state-of-the-art online multi-label technique. This section discusses the results obtained by the proposed method and compares it with the existing methods. The results obtained from the proposed method are evaluated for consistency, performance, and speed.

### 5.1 Consistency

Consistency is a key feature that is essential for any new technique proposed. Any technique proposed should provide consistent results for multiple trials with minimal variance. The consistency of a technique can be identified using cross-validation procedure. Therefore, a 5-fold cross validation and a 10-fold cross validation is performed on the proposed technique for each of the 5 datasets. Since the initial weights are assigned randomly for an ELM based technique, it is critical to evaluate the consistency of the proposed technique.

Table 2. Methods used for comparison

| Method Name | Method Category | Machine Learning Category |
|---|---|---|
| Classifier Chain (CC) | PT | SVM |
| QWeighted approach for Multi-label Learning (QWML) | PT | SVM |
| Hierarchy Of Multi-label ClassifiERs (HOMER) | PT | SVM |
| Multi-Label C4.5 (ML-C4.5) | AA | Decision Trees |
| Predictive Clustering Trees (PCT) | AA | Decision Trees |
| Multi-Label k-Nearest Neighbors (ML-kNN) | AA | Nearest Neighbors |
| Ensemble of Classifier Chains (ECC) | EN | SVM |
| Random Forest Predictive Clustering Trees (RF-PCT) | EN | Decision Trees |
| Random Forest of ML-C4.5 (RFML-C4.5) | EN | Decision Trees |

The unique feature of multi-label classification is the possibility of the partial correctness of the classifier. Therefore, calculating the error rate for multi-label problems is not same as that of traditional binary or multi-class problems. One or more of the multiple labels to which the sample instance belongs and/or the number of labels the sample instance belongs can be identified partially correctly resulting in the partial correctness of the classifier. Hence, the hamming loss performance metric is used to quantitatively measure the correctness of the classifier. The hamming loss is a measure of the misclassification rate of the learning technique. The lower the hamming loss, the better is the classification accuracy.

Hamming loss gives the percentage of wrong labels to the total number of labels. It represents the number of times the sample-label pair is misclassified (Madjarov et al. 2012).

The hamming loss for an ideal classifier is zero. The hamming loss is calculated using the following expression,

$$Hamming\ Loss\ =\ \frac{1}{N}\sum_{i=1}^{N}\frac{1}{m}\ |MLC(x_i)\Delta\ Y_i| \qquad (16)$$

where, $MLC(x_i)$ denotes the predicted output of the multi-label classifier, and $Y_i$ gives the target result to be achieved.

To evaluate the consistency of the proposed method, a 5-fold and a 10-fold cross validation of hamming loss metric is carried out for each of the 5 datasets and is tabulated.

Table 3. Consistency table – cross validation

| Dataset | Hamming Loss - 5-fcv | Hamming Loss - 10-fcv |
| --- | --- | --- |
| Yeast | 0.206 ± 0.001 | 0.206 ± 0.002 |
| Scene | 0.098 ± 0.002 | 0.098 ± 0.002 |
| Corel5k | 0.009 ± 0.000 | 0.009 ± 0.000 |
| Enron | 0.049 ± 0.001 | 0.049 ± 0.001 |
| Medical | 0.011 ± 0.001 | 0.011 ± 0.001 |

From Table 3, it can be seen that the proposed technique is consistent in its performance over repeated executions and cross validations, thus, demonstrating the consistency of the technique.

### 5.2 Performance Metrics

Due to the possibility of the partial correctness of the classifier result, one specific metric will not be sufficient to quantitatively measure the performance of a technique. Therefore, a set of quantitative performance evaluation metrics is used to validate the performance of the multi-label classifier. The performance metrics used are hamming loss, accuracy, precision, recall and $F_1$-measure. (Madjarov et al. 2012)

*Hamming Loss.* The hamming loss is a measure of misclassification rate of the learning technique. The lower the hamming loss, the better is the classification accuracy. The correctness of the classification of the learning technique can be analyzed by comparing the hamming loss metric. The definition of the hamming loss and the mathematical equation are foretold in equation 16.

*Accuracy.* Accuracy of a classifier is defined as the ratio of the total number of correctly predicted labels to the total number of labels of that sample instance. The accuracy measure can be evaluated using the following expression,

$$Accuracy = \frac{1}{N} \sum_{i=1}^{N} \left( \frac{|MLC(x_i) \cap Y_i|}{|MLC(x_i) \cup Y_i|} \right) \quad (17)$$

*Precision.* Precision is the proportion of the predicted correct labels to the total number of actual labels averaged over all instances. In other words, it is the ratio of true positives to the sum of true positives and false positives averaged over all instances. Precision can be computed as follows,

$$Precision = \frac{1}{N} \sum_{i=1}^{N} \left( \frac{|MLC(x_i) \cap Y_i|}{|MLC(x_i)|} \right) \quad (18)$$

*Recall.* Recall is the proportion of the predicted correct labels to the total number of predicted labels averaged over all instances. In other words, it is the ratio of true positives to the sum of true positives and false negatives averaged over all instances. The expression for recall is given as follows:

$$Recall = \frac{1}{N} \sum_{i=1}^{N} \left( \frac{|MLC(x_i) \cap Y_i|}{|Y_i|} \right) \quad (19)$$

*$F_1$ measure.* $F_1$ measure is given by the harmonic mean of Precision and Recall. The expression to evaluate $F_1$ measure is given by,

$$F1 - measure = \frac{1}{N} \sum_{i=1}^{N} \left( \frac{2 * |MLC(x_i) \cap Y_i|}{|MLC(x_i)| + |Y_i|} \right) \quad (19)$$

The proposed method is experimented on five different datasets for the five different performance metrics, and the results are tabulated. From Table 4, it can be seen that, the proposed method has a very low hamming loss and better performance metric measures for a wide range of datasets irrespective of the label density and label cardinality values.

Table 4. Performance metrics of OSML-ELM

| Dataset | Hamming Loss | Accuracy | Precision | Recall | F$_1$ measure |
| --- | --- | --- | --- | --- | --- |
| Yeast | 0.206 | 0.493 | 0.693 | 0.580 | 0.632 |
| Scene | 0.098 | 0.610 | 0.630 | 0.645 | 0.637 |
| Corel5k | 0.009 | 0.060 | 0.175 | 0.063 | 0.093 |
| Enron | 0.049 | 0.404 | 0.640 | 0.461 | 0.536 |
| Medical | 0.011 | 0.713 | 0.760 | 0.740 | 0.750 |

## 5.3 Comparison with State-of-the-Arts Techniques

The performance of the proposed method is compared with nine state-of-the-art techniques as specified in Table 2. Hamming loss performance metric provides the percentage of wrong labels to the total number of labels. Accuracy performance metric provides the ratio of the total number of correctly predicted labels to the total number of labels of that sample instance. Therefore, hamming loss and accuracy are the key performance metrics for evaluating the performance of the proposed method. The hamming loss and accuracy metrics are used to compare the performance of the proposed technique with the state-of-the-art techniques. The comparison results are given in Fig 2 and Fig 3 respectively.

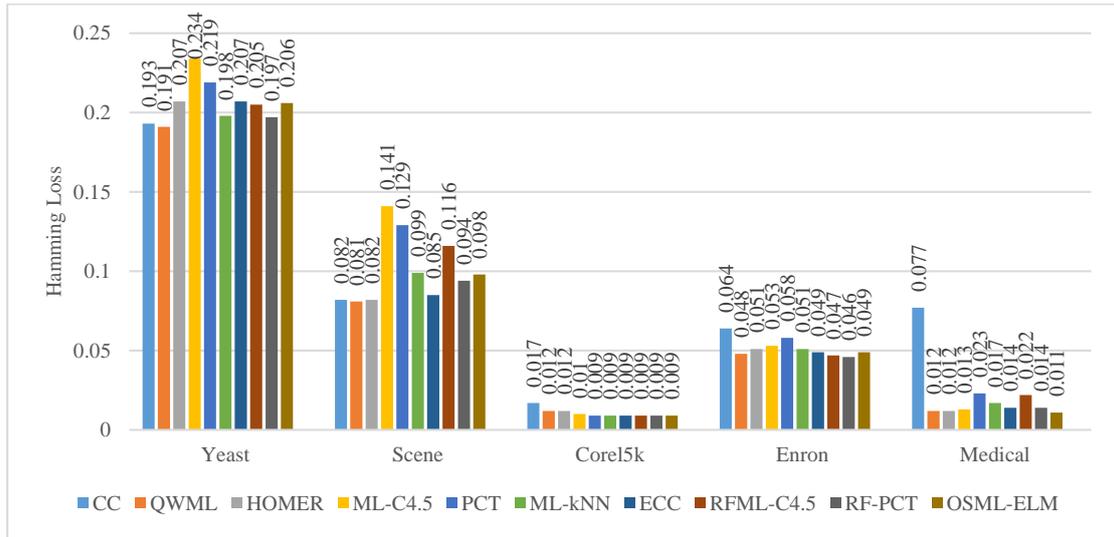

Fig. 2. Comparison of hamming loss metric with state-of-the-art techniques

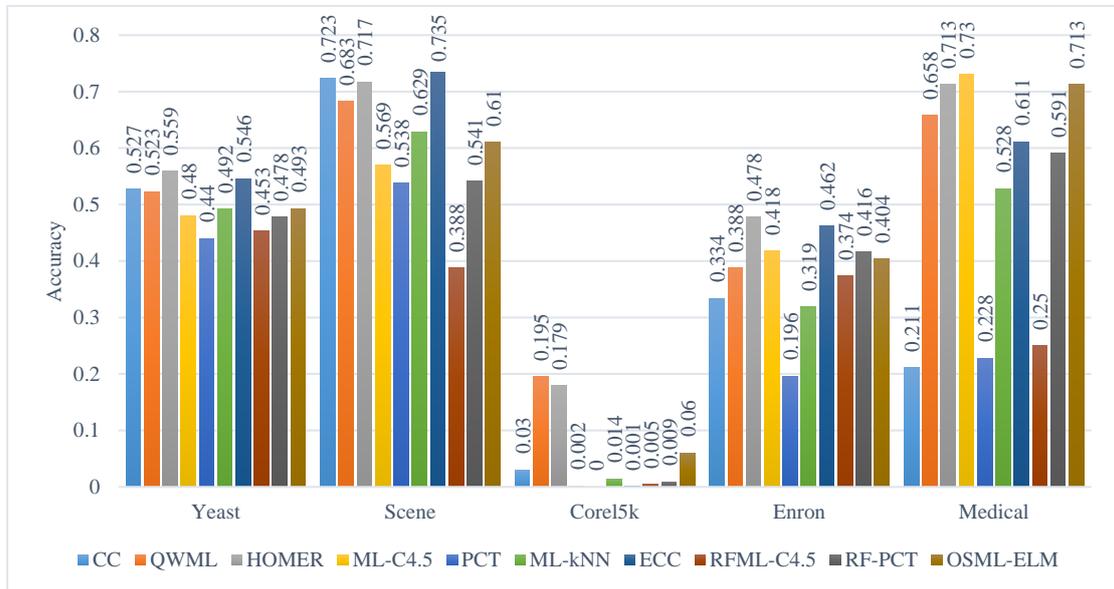

Fig. 3. Comparison of accuracy metric with state-of-the-art techniques

Hamming loss is the measure of misclassification in the dataset. Lower the hamming loss, better the performance of the classifier. For an ideal classifier, the hamming loss is equal to zero. Accuracy is the ratio of number of correctly predicted labels to the total number of labels for a given sample. Higher the accuracy, better the performance of the classifier. It is evident from the figure that, among the 10 different multi-label classifiers, the proposed method ranks among the top methods for all datasets, thus, outperforming most of the existing state-of-the-art techniques.

## 5.4 Comparison of Execution Speed

The performance of the proposed method in terms of execution speed is evaluated by comparing the training time and the testing time of the algorithm used. The proposed method is applied to 5 different datasets from various application domains and a wide range of label density and label cardinality values. The comparison of training time and testing time of the proposed method with existing state-of-the-art methods are tabulated in Tables 5 and 6.

Table 5. Training time comparison

| Dataset | CC | QWML | HOMER | ML-C4.5 | PCT | ML-kNN | ECC | RFML-C4.5 | RF-PCT | OSML-ELM |
|---|---|---|---|---|---|---|---|---|---|---|
| **Yeast** | 206 | 672 | 101 | 14 | 1.5 | 8.2 | 497 | 19 | 25 | **0.114** |
| **Scene** | 99 | 195 | 68 | 8 | 2 | 14 | 319 | 10 | 23 | **2.329** |
| **Corel 5k** | 1225 | 2388 | 771 | 369 | 30 | 389 | 10073 | 385 | 902 | **5.365** |
| **Enron** | 440 | 971 | 158 | 15 | 1.1 | 6 | 1467 | 25 | 47 | **0.630** |
| **Medical** | 28 | 40 | 16 | 3 | 0.6 | 1 | 103 | 7 | 27 | **0.663** |

Table 6. Testing time comparison

| Dataset | CC | QWML | HOMER | ML-C4.5 | PCT | ML-kNN | ECC | RFML-C4.5 | RF-PCT | OSML-ELM |
|---|---|---|---|---|---|---|---|---|---|---|
| **Yeast** | 25 | 64 | 17 | 0.1 | 0 | 5 | 158 | 0.5 | 0.2 | **0.017** |
| **Scene** | 25 | 40 | 21 | 1 | 0 | 14 | 168 | 2 | 1 | **0.047** |
| **Corel 5k** | 31 | 119 | 14 | 1 | 1 | 45 | 2077 | 1.8 | 2.5 | **0.076** |
| **Enron** | 53 | 174 | 22 | 0.2 | 0 | 3 | 696 | 1 | 1 | **0.028** |
| **Medical** | 6 | 25 | 1.5 | 0.1 | 0 | 0.2 | 46 | 0.5 | 0.5 | **0.039** |

From Tables 5 and 6, it can be clearly seen that the proposed OSML-ELM outperforms all the existing techniques in terms of execution speed. Despite being an online learning algorithm, the speed of the proposed OSML-ELM is several folds faster than most of the existing batch learning techniques. This high speed nature of the OSML-ELM will enable it to perform real-time multi-label classification on streaming data. It is to be highlighted that there are no existing techniques in the literature that can perform real-time online multi-label classification.

## 5.5 Real-Time Classification

For an online classifier to perform classification in real-time, the time taken for executing a single block of data (epoch) should be very low. If the time taken for processing an epoch is more than the rate of arrival of the sequential data, real-time processing of the streaming data cannot be achieved. From the results obtained for the training time of the classifier, the average time required for the execution of a single block of data can be estimated. The number of epochs is identified by the number of times the sequential learning phase is executed while experimenting with the specific dataset. The average time of execution to process a single block of data for the five different datasets are tabulated.

Table 7. Average time per epoch

| **Dataset** | **Training Time (s)** | **Number of epochs** | **Average time(s)/epoch** |
| --- | --- | --- | --- |
| **Yeast** | 0.114 | 51 | 0.00223529 |
| **Scene** | 2.329 | 48 | 0.04852083 |
| **Corel5k** | 5.365 | 93 | 0.05768817 |
| **Enron** | 0.63 | 48 | 0.013125 |
| **Medical** | 0.663 | 37 | 0.01791892 |

From Table 7, it can be seen that the proposed OSML-ELM can perform multi-label classification for streaming data applications with high accuracy and high speed. Also, the proposed method is compared with the state-of-the-art online multi-label technique by (Hua and Qi 2008). Since the active learning technique is multimedia specific, scene dataset is used to compare the performance. The paper lists the $F_1$ score of the active learning method for scene dataset to be 0.5612. The proposed method achieves a $F_1$ score of 0.6371 for the same scene dataset and is achieved in real-time streaming data. This shows that the proposed method has performed better in terms of speed and performance over the existing multi-label classifiers. The key advantage of the proposed method is that OSML-ELM is capable of performing multi-label classification in real time. It is to be highlighted that there are no online multi-label classifiers that can perform the multi-label classification in real time.

## 6. Conclusions

The proposed OSML-ELM classifier outperforms the existing state-of-the-arts multi-label classification techniques in terms of speed and performance. The application areas of multi-label classification are rapidly increasing due to its generality and several real-world applications require the need for multi-label classification. Due to its increased complexity and wide variations in the characteristics of multi-label datasets based on label density and label cardinality, a classifier that performs well for one dataset might not perform well for a different dataset. Also, high-speed real-time classification of multi-label data is required for real-world applications. The OSML-ELM method is a novel generic real-time multi-label classifier that performs uniformly well on datasets of a wide range of label density, label cardinality and application domains. The performance of the proposed method is compared with five datasets and nine different state-of-the-arts techniques. It can be seen from the results that the proposed OSML-ELM is a key progress in achieving real-time multi-label classification for streaming data applications. The proposed OSML-ELM can be extended further to learn new data labels progressively from the data stream by retaining the previously learnt knowledge without the need for retraining.